\documentclass[runningheads]{llncs}

\usepackage[T1]{fontenc}
\usepackage{graphicx,verbatim}
\usepackage{amsmath}
\usepackage{amssymb}
\usepackage{booktabs}
\usepackage{array}
\usepackage{multirow}
\usepackage{bm}
\usepackage{url}
\usepackage[hidelinks]{hyperref}

\newcommand{\iou}{\mathrm{IoU}}
\newcommand{\monitor}{TCSR-Monitor}
\newcommand{\unet}{U-Net}

\begin{document}

\title{Beyond Uncertainty: Generalizable Failure Monitoring for Surgical Segmentation under Acquisition Degradation}
\titlerunning{Monitoring Surgical Segmentation Failures}

\author{Hieu D. Pham \and Dang P. M. Cao \and Thanh Trung Huynh}
\authorrunning{H. D. Pham et al.}
\institute{College of Engineering and Computer Science, VinUniversity, Hanoi, Vietnam \\
    \email{24hieu.pd@vinuni.edu.vn}}

\maketitle

\begin{abstract}
Surgical segmentation networks can fail silently under acquisition degradation: predicted masks may be wrong even when model confidence remains high. Existing deployment-time monitors rely primarily on uncertainty estimates and can therefore miss confident failures. We present \monitor{} (\textbf{T}emporal \textbf{C}onformal \textbf{S}urgical \textbf{R}isk Monitor), a post-hoc failure-monitoring framework that combines confidence with observable shape, temporal-consistency, and image-quality cues. \monitor{} wraps a frozen segmentation model, requires no model internals, and operates without ground truth at deployment. We also introduce a validation protocol to assess whether alarms remain credible under distribution shift. On EndoVis~2017, leave-one-corruption-out evaluation shows that \monitor{} generalizes to unseen acquisition degradations and substantially outperforms confidence-based baselines. A circularity control confirms that it predicts segmentation failure rather than simply detecting corrupted images. Mondrian conformal calibration balances miss-rates across degradation severities, but a single global threshold still produces false alarms on up to 40\% of correctly segmented frames at moderate corruption. Zero-shot transfer to SAM2 demonstrates feature portability, although entropy outperforms the transferred monitor at both evaluated thresholds. Overall, reliable monitoring under acquisition degradation benefits from complementary observable signals beyond confidence alone, but substantial false-alarm and transfer limitations remain. Code and trained configurations are available at \href{https://github.com/dinhieufam/tcsr-monitor}{\texttt{github.com/dinhieufam/tcsr-monitor}}.

\end{abstract}

\section{Introduction}

Deep segmentation networks are increasingly used to parse endoscopic video into
instrument and anatomy masks~\cite{ref:shvets,ref:allan}, and their output
increasingly feeds downstream tasks such as tool tracking, skill assessment, and
intra-operative guidance. Deployed surgical video, however, contains smoke,
defocus, illumination change, compression, and motion blur that are
under-represented in the curated benchmarks on which these networks are tuned.
Such acquisition shifts can produce \emph{silent failures}: masks with low
intersection-over-union (IoU) that the network nonetheless reports with high
confidence. These failures are the most dangerous kind, because a confident but
incorrect mask never trips a confidence-based alarm and can propagate unnoticed
into whatever consumes it.

Most deployment-time monitoring methods estimate uncertainty through predictive
entropy~\cite{ref:malinin}, max-softmax confidence~\cite{ref:hendrycks}, or
calibration-based scores~\cite{ref:guo}. Such signals are effective when
prediction errors coincide with high uncertainty, but they become unreliable
precisely when a segmenter produces a confident yet incorrect mask. Acquisition
degradation is a common source of this mismatch: it pushes the input away from
the training distribution while leaving the output logits sharp, so uncertainty
stays low even as segmentation quality collapses. Deployment-time monitoring
therefore cannot rely on uncertainty alone, and needs signals that remain
informative when the network is confidently wrong.

Learned failure predictors show that auxiliary signals can recover some of this
missing information~\cite{ref:corbiere,ref:rottmann,ref:fsnet}, but two questions
remain open for surgical deployment. First, it is unclear whether such monitors
generalize beyond the degradation types seen during training, or whether they
merely fingerprint the specific corruption operators used to build the training
set. Second, many evaluations cannot separate genuine failure prediction from
the far easier task of recognizing that an image is degraded at all. A monitor
that only detects corruption, or only recognizes familiar corruptions, offers
little protection in the field.

We address these gaps with \monitor{}, a post-hoc failure monitor that combines
multiple observable cues---output confidence, mask geometry, temporal
consistency, and image quality---into a single segmentation-failure-risk score.
The monitor wraps a frozen segmentation model, requires no access to model
internals or gradients, and operates without ground truth during deployment,
which keeps it compatible with clinical models that cannot be retrained or
instrumented. More importantly, we argue that a learned failure monitor is meaningful only if
its alarms stay trustworthy under distribution shift, and we treat that
credibility as a claim to be tested rather than assumed. We therefore evaluate
\monitor{} with three complementary protocols: generalization to unseen
acquisition degradations, discrimination of segmentation failure from mere image
corruption, and calibrated safety across degradation severity. Together they ask
not only whether the monitor scores well, but whether its score means what a
deploying clinician would need it to mean. We make the following contributions:

\begin{enumerate}
\item We propose a deployment-time failure-monitoring framework that integrates
complementary observable cues beyond uncertainty-based confidence scores.
\item We introduce a validation protocol---leave-one-corruption-out (LOCO)
generalization, a circularity control, and severity-aware calibration---that
tests whether such a monitor stays credible under distribution shift.
\item We show that uncertainty-based confidence alone is insufficient under
acquisition degradation, whereas complementary observable cues improve
robustness to unseen corruptions.
\item We use Mondrian conformal calibration to balance miss-rates across
degradation severities while making the resulting false-alarm cost explicit.
\item We provide supporting controls in the supplement---confidence-only,
learner, temporal-order, sequence-held-out, and SAM2 transfer---that bound
rather than expand these claims.
\end{enumerate}

The intended use is not to replace segmentation, but to add a fast, external
watchdog for cases in which the segmenter remains confident while its mask
quality collapses. This distinction also determines the evaluation: a monitor
that only succeeds on degradation types observed during training provides
limited deployment value, so practical monitoring must generalize to previously
unseen acquisition conditions.

%%%%%%%%%%%%%%%%%%%%%%%%%%%%%%%%%%%%%%%%%%%%%%%%%%%%%%%%%%%%%%%%%%%%%%%%%%%%%%%%
\section{Related Work}

Instrument segmentation has advanced through \unet{} variants~\cite{ref:ronneberger},
transformers~\cite{ref:dong}, and foundation models such as SAM/SAM2
~\cite{ref:kirillov,ref:ravi}. Robustness benchmarks expose sensitivity to
common or surgical corruptions~\cite{ref:hendrycks_c,ref:segstrong,ref:michieli},
but they primarily evaluate the segmenter itself. Our focus is deployment-time
monitoring after a mask has already been produced.

Confidence scores, entropy, calibration, test-time ensembles, and feature-space
OOD scores are common post-hoc reliability signals
~\cite{ref:hendrycks,ref:malinin,ref:guo,ref:lakshmi,ref:lee}. In medical
segmentation, however, uncertainty can be calibrated in aggregate while still
failing on individual cases~\cite{ref:jungo}, and networks may remain
overconfident on wrong masks~\cite{ref:mehrtash}. Learned failure prediction
therefore provides a complementary route: prior work estimates model confidence
or segment quality from auxiliary signals~\cite{ref:corbiere,ref:granese,ref:rottmann,ref:fsnet}.
We differ by centering surgical acquisition degradation, held-out corruption
protocols, and a circularity control that separates failure prediction from
corruption detection. Our alarm layer relates to selective prediction
~\cite{ref:geifman,ref:selectivenet} and conformal risk control
~\cite{ref:vovk,ref:angelo_intro,ref:crc}; group-conditional Mondrian
calibration~\cite{ref:mondrian} is especially natural when degradation severity
defines safety-relevant groups.

%%%%%%%%%%%%%%%%%%%%%%%%%%%%%%%%%%%%%%%%%%%%%%%%%%%%%%%%%%%%%%%%%%%%%%%%%%%%%%%%
\section{Method}

\begin{figure}[t]
\centering
\includegraphics[width=\textwidth]{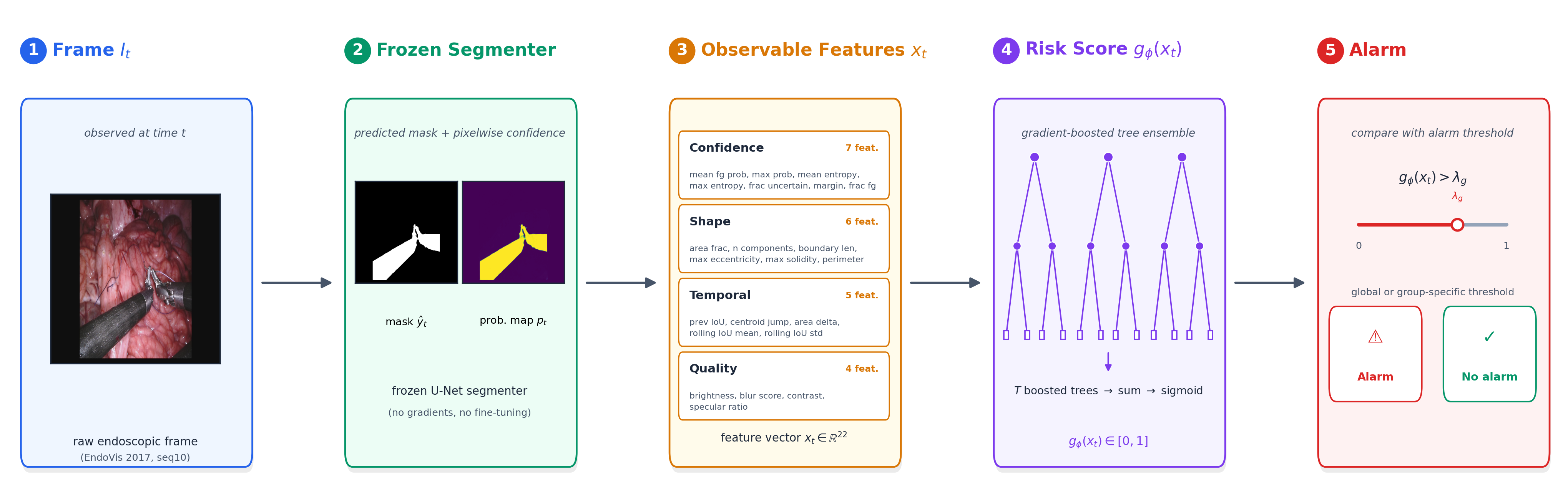}
\caption{Overview of \monitor{}. The monitor observes the raw frame and the
segmenter's output, but does not modify the segmenter or use ground truth at
deployment. Calibration converts risk scores into global or group-specific
alarm thresholds.}
\label{fig:pipeline}
\end{figure}

\monitor{} wraps a frozen segmenter $f_\theta$ without changing its weights
(Fig.~\ref{fig:pipeline}). For each frame $I_t$, the segmenter produces a
probability map $p_t$ and mask $\hat y_t$. A frame is labeled a failure during
training/evaluation when
\begin{equation}
\mathrm{fail}(I_t)=\mathbf{1}\!\left[\iou(\hat{y}_t,y_t^\star)<\tau\right],
\qquad \tau \in \{0.5,\,0.75\}.
\label{eq:failure_label}
\end{equation}

\noindent The monitor extracts $22$ scalar features from $p_t$, $\hat y_t$,
$I_t$, and previous masks: confidence statistics, mask morphology, temporal
consistency, and image-quality measures (brightness, blur, contrast, specular
ratio). Features require only the raw frame and segmenter output, and take
under $0.5$\,ms per frame (median $0.48$\,ms, $p_{95}$ $0.53$\,ms).

The feature design is intentionally restricted to quantities available outside
the segmentation model. Confidence features capture whether the output looks
uncertain; shape features capture implausible mask geometry and fragmentation;
temporal features capture abrupt changes across adjacent video frames; and
image-quality features capture acquisition conditions that can make a confident
mask suspect. This makes the monitor compatible with frozen clinical models
and avoids relying on architecture-specific internal activations. Table
~\ref{tab:features} lists all $22$ features with their exact definitions.

\begin{table}[t]
\caption{The full set of $22$ observable features. $p_t$ is the pixelwise
probability map, $\hat y_t=\mathbf{1}[p_t\ge0.5]$ the predicted mask, and
$t\!-\!1$ the previous frame in the same video.}
\label{tab:features}
\centering
\scriptsize
\setlength{\tabcolsep}{4pt}
\begin{tabular}{lll}
\toprule
Group & Feature & Definition \\
\midrule
\multirow{7}{*}{Confidence (7)}
 & \texttt{conf\_mean\_fg\_prob} & Mean $p_t$ over foreground pixels ($\hat y_t=1$) \\
 & \texttt{conf\_max\_prob}      & $\max p_t$ over the frame \\
 & \texttt{conf\_mean\_entropy}  & Mean per-pixel binary entropy of $p_t$ \\
 & \texttt{conf\_max\_entropy}   & Max per-pixel binary entropy of $p_t$ \\
 & \texttt{conf\_frac\_uncertain}& Fraction of pixels with entropy $>0.9\log 2$ \\
 & \texttt{conf\_margin}         & Mean $|p_t-0.5|$ over the frame \\
 & \texttt{conf\_frac\_fg}       & Foreground area fraction of $\hat y_t$ \\
\midrule
\multirow{6}{*}{Shape (6)}
 & \texttt{shape\_area\_frac}       & Foreground pixel fraction of $\hat y_t$ \\
 & \texttt{shape\_n\_components}    & Number of connected foreground components \\
 & \texttt{shape\_boundary\_len}    & Total component perimeter / frame area \\
 & \texttt{shape\_max\_eccentricity}& Eccentricity of the largest component \\
 & \texttt{shape\_max\_solidity}    & Solidity (area / convex-hull area), largest component \\
 & \texttt{shape\_perimeter}        & Total (unnormalized) perimeter, summed over components \\
\midrule
\multirow{5}{*}{Temporal (5)}
 & \texttt{temp\_prev\_iou}          & $\iou(\hat y_t,\hat y_{t-1})$ \\
 & \texttt{temp\_centroid\_jump}     & Normalized centroid displacement vs.\ $t\!-\!1$ (largest component) \\
 & \texttt{temp\_area\_delta}        & $|\text{area}(\hat y_t)-\text{area}(\hat y_{t-1})|$ \\
 & \texttt{temp\_rolling\_iou\_mean} & Mean frame-to-frame IoU over a trailing $5$-frame window \\
 & \texttt{temp\_rolling\_iou\_std}  & Std. of frame-to-frame IoU over the same window \\
\midrule
\multirow{4}{*}{Quality (4)}
 & \texttt{qual\_brightness}     & Mean grayscale intensity of $I_t$, normalized to $[0,1]$ \\
 & \texttt{qual\_blur\_score}    & Variance of the Laplacian of $I_t$ (sharpness) \\
 & \texttt{qual\_contrast}       & Std.\ of grayscale intensity of $I_t$, normalized to $[0,1]$ \\
 & \texttt{qual\_specular\_ratio}& Fraction of pixels with HSV value $>240$ (specular highlight) \\
\bottomrule
\end{tabular}
\end{table}

The failure-risk model $g_\phi(x_t)\in[0,1]$ is an XGBoost classifier
~\cite{ref:xgboost}; learner ablations in the supplement show that the
representation is not unique to XGBoost. Training uses clean out-of-fold
predictions and corrupted training frames, while threshold calibration is held
out from model fitting. At deployment, ground truth is unavailable and an alarm
is raised when $g_\phi(x_t)>\lambda$.

The safety objective is miss-rate control: among failed frames, the fraction
not alarmed should stay below target $\alpha$:
\begin{equation}
\mathrm{miss}(\lambda)=
\frac{\bigl|\{\, i : g_\phi(x_i) \le \lambda,\; \mathrm{fail}(I_i)=1\,\}\bigr|}
     {\bigl|\{\, i : \mathrm{fail}(I_i)=1 \,\}\bigr|}.
\end{equation}
Global split-conformal calibration~\cite{ref:vovk} chooses one threshold
$\lambda$ on a calibration split. Because acquisition severity changes the
failure distribution, we also use Mondrian calibration~\cite{ref:mondrian}:
frames are partitioned into groups $g$ (corruption severity for degraded
frames; blur tertile for clean frames), and each group receives its own
threshold
\begin{equation}
\lambda_g = \sup\!\left\{ \lambda :
\frac{\bigl|\{\, i \in \mathcal{G}_g : g_\phi(x_i) \le \lambda,\; y_i = 1 \,\}\bigr|}
     {\bigl|\{\, i \in \mathcal{G}_g : y_i = 1 \,\}\bigr|} \le \alpha \right\}.
\end{equation}
This equalizes empirical conservatism across groups rather than letting easy
conditions dominate the average.

The thresholds are calibrated after score learning. Thus discrimination
(whether failures rank above correct frames) and safety control (where to set
the alarm threshold) are evaluated separately. This separation is important in
rare-event settings: a high AUROC monitor can still be operationally burdensome
if the threshold required to catch failures creates many frame-level alarms.

%%%%%%%%%%%%%%%%%%%%%%%%%%%%%%%%%%%%%%%%%%%%%%%%%%%%%%%%%%%%%%%%%%%%%%%%%%%%%%%%
\section{Experiments}

\subsection{Setup}

We use EndoVis~2017 Instrument Segmentation~\cite{ref:bodenstedt}: ten robotic
surgical sequences, with sequences $1$--$7$ for train/calibration and
$8$--$10$ for test ($1575/450/825$ frames). The frozen segmenter is a
ResNet34-\unet{} fine-tuned on the training sequences. We apply six corruption
types at five severity levels (Gaussian blur/noise, motion blur, brightness,
contrast, JPEG), yielding $24{,}750$ corrupted test frames. We evaluate
$\tau=0.5$ and $\tau=0.75$; clean-test failure prevalence is $1.7\%$ and
$6.9\%$, respectively. Baselines are Max-Softmax, Entropy, and a Temporal
Heuristic---risk score $1-\iou(\hat y_t,\hat y_{t-1})$ between consecutive
predicted masks, so large frame-to-frame mask changes are flagged as risky
(score $0$ on the first frame of a sequence, where no previous mask exists)---plus
a learned confidence-only monitor in the supplement.

We use four generalization protocols. \emph{Zero-shot} trains the monitor on
clean frames only and tests on all corruptions. \emph{LOCO} trains on five
corruption types and tests on the held-out sixth type; this is the robustness
headline because the test degradation operator is unseen. \emph{Severity
extrapolation} trains on severities $1$--$3$ and tests on severities $4$--$5$.
\emph{In-distribution} trains and tests across all corruption types and is
reported only as an upper bound.

\subsection{Clean Test and Generalization}

On held-out clean test (Table~\ref{tab:main}), \monitor{} improves AUROC over
all audited baselines at both thresholds. AUPRC is prevalence-sensitive:
entropy is stronger at $\tau=0.5$, whereas \monitor{} gives the best audited
AUPRC at $\tau=0.75$.

\begin{table}[t]
\caption{Held-out clean test results (seed~0; point estimates). AUPRC should
be interpreted relative to the rare failure prevalence. Frame-level point
estimates on this split are correlated within only $3$ test sequences; see the
sequence-level bootstrap discussion below for uncertainty bounds.}
\label{tab:main}
\centering
\footnotesize
\setlength{\tabcolsep}{4pt}
\resizebox{\textwidth}{!}{%
\begin{tabular}{lcccc}
\toprule
Method & $\tau=0.5$ AUROC & $\tau=0.5$ AUPRC & $\tau=0.75$ AUROC & $\tau=0.75$ AUPRC \\
\midrule
\textbf{\monitor{} (ours)} & \textbf{0.877} & 0.468 & \textbf{0.793} & \textbf{0.301} \\
Entropy~\cite{ref:malinin}            & 0.764 & \textbf{0.635} & 0.483 & 0.220 \\
Max-Softmax~\cite{ref:hendrycks}      & 0.536 & 0.087 & 0.509 & 0.085 \\
Temporal Heuristic                    & 0.338 & 0.013 & 0.460 & 0.074 \\
\midrule
Failure prevalence                    & \multicolumn{2}{c}{$1.7\%$ ($14/825$)} & \multicolumn{2}{c}{$6.9\%$ ($57/825$)} \\
\bottomrule
\end{tabular}
}
\end{table}

\begin{figure}[t]
  \centering
  \includegraphics[width=\textwidth]{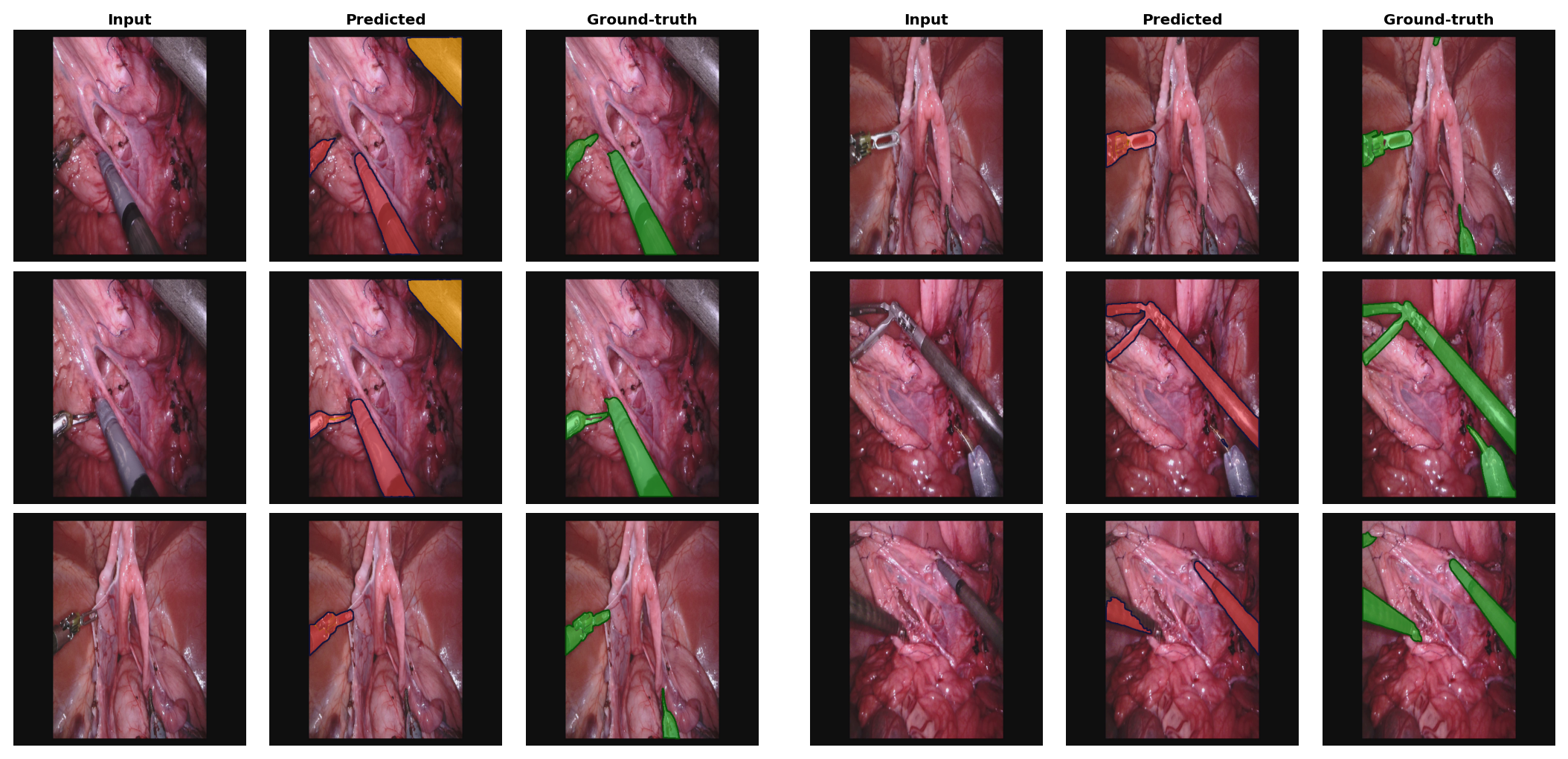}
  \caption{Qualitative examples of sub-threshold confident failures
  ($0.5 \le \iou < 0.75$). Each triplet shows the input, predicted mask, and
  ground-truth mask. The segmenter produces plausible-looking but incomplete
  or over-extended instruments, illustrating why an external failure monitor is
  needed even when the output appears confident.}
  \label{fig:qualitative}
\end{figure}

The $825$ test frames come from only $3$ surgical sequences (seq08/09/10) and
are strongly correlated within a sequence, so a frame-level bootstrap would
understate uncertainty. We instead resample the $3$ sequences with replacement
($5000$ resamples; at $\tau=0.5$ seq09 contributes zero failures, so $195/5000$
degenerate resamples are excluded) and recompute each metric on the pooled
resample. The resulting $95\%$ intervals for \monitor{} AUROC are
$[0.48,0.99]$ at $\tau=0.5$ and $[0.69,0.99]$ at $\tau=0.75$ (entropy:
$[0.41,0.90]$ and $[0.07,0.90]$). These are wide because they reflect only $3$
independent clusters, not an unstable estimator; we report the honest
sequence-level interval rather than a narrower, misleading frame-level one.
Fig.~\ref{fig:qualitative} shows the sub-threshold cases driving the
$\tau=0.75$ positive class.

\begin{table}[t]
\caption{Generalization hierarchy and LOCO breakdown (AUROC). LOCO is the
robustness headline; in-distribution is an upper bound. \emph{Wins} counts the
(corruption type, severity) cells in which \monitor{} AUROC exceeds entropy
AUROC, out of $30$ cells ($6$ corruption types $\times$ $5$ severities) for
zero-shot/LOCO/in-distribution, and $12$ cells ($6$ types $\times$ severities
$4$--$5$) for severity extrapolation; per-corruption rows report wins out of
$5$ severities.}
\label{tab:hierarchy}
\centering
\footnotesize
\setlength{\tabcolsep}{4pt}
\begin{tabular}{lccc}
\toprule
Protocol / held-out type & Monitor & Entropy & Wins \\
\midrule
Zero-shot & 0.753 & 0.465 & 24/30 \\
\textbf{LOCO overall} & \textbf{0.814} & \textbf{0.481} & \textbf{26/30} \\
Severity extrapolation & 0.831 & 0.408 & 8/12 \\
In-distribution upper bound & 0.827 & 0.465 & 26/30 \\
\midrule
Brightness        & 0.801 & 0.331 & 4/5 \\
Contrast          & 0.912 & 0.839 & 3/5 \\
Gaussian blur     & 0.806 & 0.455 & 5/5 \\
Gaussian noise    & 0.811 & 0.319 & 4/5 \\
JPEG compression  & 0.773 & 0.444 & 5/5 \\
Motion blur       & 0.779 & 0.497 & 5/5 \\
\bottomrule
\end{tabular}
\end{table}

LOCO ($0.814$, Table~\ref{tab:hierarchy}) is the robustness headline because
the test corruption type is absent from training. The small gap to the
in-distribution upper bound ($0.827$) argues against simple corruption
fingerprinting, and the monitor beats entropy on all six held-out types.

The clean-test and LOCO results answer different questions. Clean test measures
behavior on nominal deployment-like frames, where failures are rare and AUPRC
is prevalence-sensitive. LOCO stresses whether the learned score captures
transferable failure patterns under acquisition degradation. The combination is
more informative than either setting alone: clean test guards against excessive
alarms on normal video, while LOCO guards against memorizing the training
corruptions.

\subsection{Circularity and Conformal Safety}

To test whether \monitor{} predicts \emph{failure} rather than simply
\emph{corruption}, we evaluate only corrupted test frames and discriminate
correct masks ($\iou\ge0.75$) from failed masks ($\iou<0.75$). Table
~\ref{tab:circularity} shows AUROC $0.946$, while entropy is only $0.594$.
This control is essential because a monitor could otherwise appear successful
by detecting low image quality rather than segmentation failure. Restricting
the evaluation to corrupted frames removes the easy corrupted-versus-clean
shortcut and asks whether the monitor can tell when the segmenter is actually
wrong within degraded video.

\begin{table}[t]
\caption{Circularity control and Mondrian conformal calibration. Left:
within-corrupted AUROC for failure vs.\ correct masks. Right: miss-rate by
severity at $\alpha=0.10$.}
\label{tab:circularity}
\centering
\footnotesize
\begin{tabular}{lccc@{\qquad}lcc}
\toprule
Severity & Monitor & Entropy & FA-correct & Severity & Global miss & Mondrian miss \\
\midrule
1 & 0.807 & 0.484 & 8.0\% & 1 & 50.5\% & 9.3\% \\
2 & 0.822 & 0.505 & 22.3\% & 2 & 22.5\% & 10.1\% \\
3 & 0.827 & 0.483 & 40.4\% & 3 & 9.2\% & 9.9\% \\
4 & 0.861 & 0.366 & 35.8\% & 4 & 6.1\% & 10.0\% \\
5 & 0.825 & 0.478 & 8.9\% & 5 & 5.2\% & 9.9\% \\
\midrule
\textbf{Overall} & \textbf{0.946} & \textbf{0.594} & \textbf{12.6\%} &
\textbf{Overall} & \textbf{10.0\%} & \textbf{9.9\%} \\
\bottomrule
\end{tabular}
\end{table}

The FA-correct column deserves explicit emphasis, not just disclosure: at the
single global threshold used for the circularity control, the monitor
false-alarms on up to $40.4\%$ of correctly segmented frames at severity~3
(rising from $8.0\%$ at severity~1, then falling to $35.8\%$ and $8.9\%$ at
severities~4--5 as failures themselves become common). Flagging two in five
correct masks at moderate corruption is not yet a usable clinical alarm
stream at a single global threshold; Section~\ref{sec:discussion} returns to
this alongside the SAM2 result below, where uncertainty is the stronger
baseline.

Mondrian calibration fixes the safety imbalance hidden by global calibration:
the same overall $10.0\%$ global miss-rate is obtained while missing $50.5\%$
of severity-1 failures. Per-severity thresholds restore near-$10\%$ miss-rate
for all severities, but at a cost: corrupted false-alarm rate increases from
$17.6\%$ to $34.2\%$ overall and from $12.8\%$ to $58.7\%$ at severity~1
(supplement). Thus the claim is safety redistribution, not clinical alarm
readiness.

\subsection{SAM2 Transfer}

We apply the \unet{}-trained monitor to SAM2-Tiny~\cite{ref:ravi} run
zero-shot with oracle bounding-box prompts. This experiment carries the
paper's clearest limitation: as Table~\ref{tab:transfer} shows, entropy
\emph{outperforms} the transferred \monitor{} at both thresholds ($0.984$ vs.\
$0.896$ at $\tau=0.5$; $0.988$ vs.\ $0.930$ at $\tau=0.75$), so the ``beyond
uncertainty'' framing does not hold uniformly---when a segmenter's failures
already manifest as high predictive uncertainty, as SAM2's zero-shot
instrument masks do, uncertainty alone is the stronger, simpler baseline.
SAM2 fails far more often than the fine-tuned \unet{} ($51.2\%$ vs.\ $6.9\%$
at $\tau=0.75$), and the transferred monitor still reaches AUROC $0.930$,
recovering $93.3\%$ of a SAM2-supervised upper bound; we read this as evidence
that the observable-feature representation transfers across segmenter
families, not that it dominates uncertainty-based monitoring. Feature
portability and entropy's continued strength are complementary findings, not
a single unqualified one.

\begin{table}[t]
\caption{Cross-model transfer to SAM2-Tiny. Entropy is the best-AUROC method
in this setting (bold); \monitor{}'s transfer efficiency relative to a
SAM2-supervised upper bound is reported separately.}
\label{tab:transfer}
\centering
\footnotesize
\resizebox{\textwidth}{!}{%
\begin{tabular}{lcc}
\toprule
Condition & $\tau=0.5$ AUROC & $\tau=0.75$ AUROC \\
\midrule
\unet{} monitor $\rightarrow$ SAM2 (zero-shot transfer) & 0.896 $[0.873, 0.915]$ & 0.930 $[0.912, 0.945]$ \\
\textbf{Entropy $\rightarrow$ SAM2}                              & \textbf{0.984} $[0.975, 0.990]$ & \textbf{0.988} $[0.979, 0.994]$ \\
SAM2-specific monitor (upper bound)                             & 0.997 $[0.995, 0.999]$ & 0.997 $[0.994, 0.999]$ \\
Transfer efficiency (transfer / upper bound)                    & 89.8\% & 93.3\% \\
\bottomrule
\end{tabular}
}
\end{table}

%%%%%%%%%%%%%%%%%%%%%%%%%%%%%%%%%%%%%%%%%%%%%%%%%%%%%%%%%%%%%%%%%%%%%%%%%%%%%%%%
\section{Discussion}
\label{sec:discussion}

\monitor{} is best understood as a generalization-aware framework for moving
beyond uncertainty, not as a claim that one classifier dominates all baselines.
The LOCO and circularity experiments address the two failure modes that would
most undermine this claim---fingerprinting the training corruption operators,
and merely detecting that an image is degraded rather than that the segmenter
failed on it, respectively. The confidence-only,
learner, temporal, and sequence controls in the supplement narrow the claim:
confidence cues are strong, the full feature set adds modest but positive LOCO
gains, and random forest matches or exceeds XGBoost. The novelty is therefore
the combination of observable features with protocol-qualified validation and
group-conditional safety calibration.

\paragraph{Feature importance.} To make the ``complementary observable cues''
claim checkable rather than asserted, Table~\ref{tab:featimp} reports XGBoost
gain-based feature importance for the clean-test monitor of
Table~\ref{tab:main}, aggregated into the four groups of Table~\ref{tab:features}.
Confidence is the largest single share ($57.9\%$, led by
\texttt{conf\_mean\_fg\_prob} at $36.0\%$), consistent with confidence being a
strong signal in aggregate. But the remaining $42.1\%$ is not diffuse noise:
quality contributes $20.6\%$ (led by \texttt{qual\_contrast}, $11.2\%$) and
shape $16.5\%$; temporal contributes least ($5.0\%$), matching the
supplement's temporal-order stress control, where shuffling frame order costs
only a modest LOCO AUROC. Two of the top-5 individual features are
image-quality, not confidence, features---quantitative, not just qualitative,
support that acquisition-quality and shape cues carry real weight in the
trained estimator.

\begin{table}[t]
\caption{XGBoost gain-based feature importance (clean-test monitor,
Table~\ref{tab:main}), aggregated by feature group and listing the top-5
individual features. Importances sum to $1$ over all $22$ features.}
\label{tab:featimp}
\centering
\scriptsize
\setlength{\tabcolsep}{5pt}
\begin{tabular}{lc@{\qquad}lc}
\toprule
Group & Share of importance & Top individual feature & Importance \\
\midrule
Confidence & $57.9\%$ & \texttt{conf\_mean\_fg\_prob} & $36.0\%$ \\
Quality    & $20.6\%$ & \texttt{qual\_contrast}       & $11.2\%$ \\
Shape      & $16.5\%$ & \texttt{conf\_margin}         & $5.7\%$ \\
Temporal   & $5.0\%$  & \texttt{conf\_frac\_fg}       & $5.6\%$ \\
           &          & \texttt{qual\_blur\_score}    & $5.2\%$ \\
\bottomrule
\end{tabular}
\end{table}

The supplement fills in the main controls behind this interpretation. The
learned confidence-only baseline is deliberately stronger than raw entropy or
max-softmax because it receives the same learner family as the full monitor.
The full feature set improves over it under LOCO, but only modestly, so the
claim is not that confidence is irrelevant. Instead, shape, temporal, and
image-quality features add incremental generalization signal where confident
failures are the target.

The main limitation is scope. All validated discrimination and calibration
results use EndoVis~2017 with synthetic corruptions; real multi-site surgical
video remains future work. A cached CholecSeg8k audit produced failures for
every frame and therefore could not provide AUROC/AUPRC external validation;
we traced this to a preprocessing defect, not a genuine zero-shot failure. Our
mask loader binarized instrument pixels using a placeholder class-ID set left
over from an early, unverified reading of the dataset documentation;
CholecSeg8k's actual instrument IDs (grasper, L-hook electrocautery) differ
from the placeholder, so binarizing against it yields an empty ground-truth
mask for nearly every frame ($58/60$ sampled), mechanically driving IoU toward
zero regardless of prediction quality. We report this as a labeling bug found
while preparing this revision and leave a corrected re-run to future work
rather than substitute an unverified fix now. SAM2 uses oracle bounding-box
prompts and does not
replace a second fine-tuned surgical-backbone study. Finally, conformal
coverage is empirical rather than certified under temporal dependence, and
frame-level alarm rates are too high for direct clinical use without event
aggregation or workflow suppression.

These limitations narrow the deployment claim. The present evidence supports
\monitor{} as a credible research baseline and validation protocol for
monitoring surgical segmentation under acquisition degradation. It does not yet
establish a clinically deployable alert stream, nor does it prove universal
transfer across hospitals, procedures, or fine-tuned backbone families.

%%%%%%%%%%%%%%%%%%%%%%%%%%%%%%%%%%%%%%%%%%%%%%%%%%%%%%%%%%%%%%%%%%%%%%%%%%%%%%%%
\section{Conclusion}

We presented \monitor{}, pairing a lightweight feature-based monitor with LOCO
generalization, a circularity control, and Mondrian conformal calibration into
a credible, bounded baseline for catching confident surgical-segmentation
failures that uncertainty scores miss under acquisition degradation---while
also delimiting where uncertainty baselines and alarm-burden limitations
remain important.

\newpage

%%%%%%%%%%%%%%%%%%%%%%%%%%%%%%%%%%%%%%%%%%%%%%%%%%%%%%%%%%%%%%%%%%%%%%%%%%%%%%%%

\end{document}